\documentclass[10pt,twocolumn,letterpaper]{article}

\usepackage[pagenumbers]{cvpr} 

\usepackage{graphicx}
\usepackage{amsmath}
\usepackage{amssymb}
\usepackage{booktabs}
\usepackage{adjustbox}

\usepackage[accsupp]{axessibility}  

\usepackage[pagebackref,breaklinks,colorlinks]{hyperref}

\usepackage[capitalize]{cleveref}
\crefname{section}{Sec.}{Secs.}
\Crefname{section}{Section}{Sections}
\Crefname{table}{Table}{Tables}
\crefname{table}{Tab.}{Tabs.}

\usepackage{xcolor,colortbl}
\definecolor{light}{RGB}{240,245,255}

\usepackage{multirow}

\newcommand\T{\rule{0pt}{2.6ex}}       
\newcommand\B{\rule[-1.2ex]{0pt}{0pt}} 

\def\OURS{our method\xspace}

\def\cvprPaperID{59} 
\def\confName{CVPR}
\def\confYear{2022}

\begin{document}

\title{Bootstrapped Representation Learning for Skeleton-Based Action Recognition}

\author{
  Olivier Moliner$^{1,2}$\quad Sangxia Huang$^{2}$ \quad Kalle Åström$^{1}$\\ 
  \normalsize{$^{1}$Centre for Mathematical Sciences, Lund University, Lund, Sweden}\\
  \normalsize{$^{2}$R\&D Center Lund Laboratory, Sony Europe B.V., Lund, Sweden}\\
  \small\texttt{\{olivier.moliner, karl.astrom\}@math.lth.se, sangxia.huang@sony.com}
}

\maketitle

\begin{abstract}
In this work, we study self-supervised representation learning for 3D skeleton-based action recognition.
   We extend Bootstrap Your Own Latent (BYOL) for representation learning on skeleton sequence data and
propose a new data augmentation strategy including two asymmetric transformation pipelines. 
We also introduce a multi-viewpoint sampling method that leverages multiple viewing angles of the same action captured by different cameras. 
In the semi-supervised setting, we show that the performance can be further improved by knowledge distillation from wider networks, leveraging once more the unlabeled samples.
We conduct extensive experiments on the NTU-60, NTU-120 and PKU-MMD datasets to demonstrate the performance of our proposed method.
Our method consistently outperforms the current state of the art on linear evaluation, semi-supervised and transfer learning benchmarks.
\end{abstract}


\section{Introduction}
\label{sec:intro}

Action recognition is an essential task in computer vision with many real-world applications, such as surveillance and assisted living, human-robot interaction, video retrieval and autonomous driving. 
Recent advances in sensors \cite{Shotton2011-zf} and human pose estimation methods \cite{Cao2019-pk, Sun2019-cl} make it possible to perform action recognition on skeleton data instead of RGB images, paving the way for light-weight action recognition algorithms that are robust to different background and lighting conditions \cite{Yan2018-pk, Li2019-ml, Liu2020-pl, Li2019-uf, Shi2019-wc, Shi2019-yj, Peng2020-jd, Cheng2020-wl}.
However, training these algorithms in a fully-supervised manner requires large datasets of 3D skeleton data with accurate annotations, which are time-consuming and costly to prepare.

To address this issue, self-supervised methods have been proposed to learn action representations from skeleton data without human-provided labels. The learned representations can subsequently be fine-tuned on a smaller set of labeled data to obtain an action recognition model. 
Early methods have been focusing on learning representations by solving reconstruction or prediction problems \cite{Zheng2018-el, Su2020-st, Kundu2018-ep}, while more recent works adopt a contrastive learning framework \cite{Lin2020-rf, Rao2021-wk, Li2021-rh}.

Contrastive learning has been used recently to reach remarkable performance in the image domain and is closing the performance gap with supervised pre-training when transferred to downstream tasks \cite{Van_den_Oord2018-ms,Misra2020-tg,Chen2020-wj,He2020-nt,Chen2020-bo,Chen2020-tw}.
These methods learn representations by mapping different augmented views of the same input sample (positive pairs) closer while pushing augmented views of different input samples (negative pairs) apart.
Contrastive methods require a large number of negative samples to achieve good performance, necessitating large batch sizes \cite{Chen2020-wj, Chen2020-bo} or memory banks of negative samples \cite{He2020-nt, Chen2020-tw}.
More recently, Bootstrap Your Own Latent (BYOL) \cite{Grill2020-ey} showed that explicit negative pairs are not required to learn transferable visual representations. Instead, it uses two networks -- an online network and a target network -- to encode two augmented views of the same input sample. The online network is trained to predict the output of the target network, and the target network is updated with an exponential moving average of the online network.

\begin{figure*} 
\centering
\includegraphics[width=\linewidth]{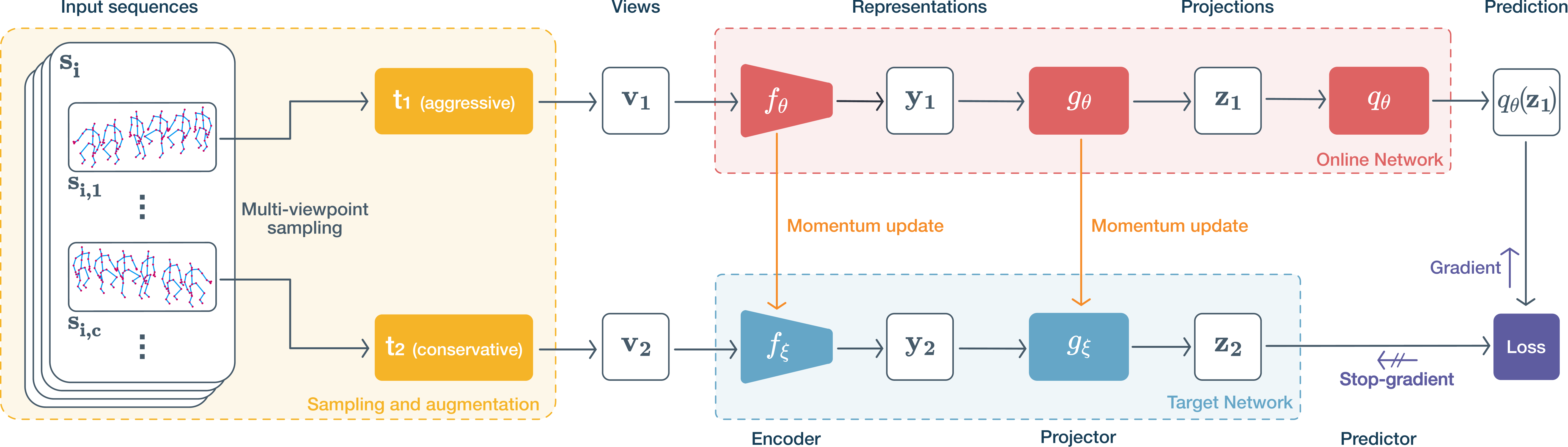}
\caption{\label{fig:byol}\textbf{Overview of our proposed method.}
We sample two random viewpoints $\mathbf{s_{i,c_1}}$, $\mathbf{s_{i,c_2}}$ 
of the same action sequence captured from different viewing angles and transform them with two asymmetric augmentation pipelines. 
The resulting views are passed to the online and target networks.
The online network is trained to predict the output of the target network.
The gradients are only propagated through the online network, while the target network's parameters are updated with an exponential moving average of the online network's parameters.
 }
\vspace{-5pt}
\end{figure*}

In this work, we adapt BYOL for skeleton-based action recognition. 
As is the case with contrastive learning, data augmentation
is an essential part of this method as it guides the network to learn relevant features in the absence of labels.
We propose new data augmentation approaches for action sequences tailored to make the learned representations robust to semantically-irrelevant variations.
We introduce two asymmetric augmentation pipelines to reduce the distribution shift between self-supervised pre-training and supervised fine-tuning.
We also propose a multi-viewpoint sampling method to leverage readily available positive pairs to produce distinct views of the same sample.
Several datasets feature the same action sequence captured simultaneously from different angles by different cameras. A naive adaptation of BYOL would treat these recordings of the 
same action sequence as unrelated samples. 
Under multi-viewpoint sampling, these are treated as positive pairs, which encourages learning view-invariant representations.

We evaluate the learned representations on linear evaluation, semi-supervised learning and transfer learning tasks.
We show that representations learned with our method outperform the current state of the art on all benchmarks.
Inspired by \cite{Chen2020-bo}, we achieve further performance improvement at the semi-supervised tasks using knowledge distillation from wider networks.

To summarize, the main contributions of this paper are as follows:
\begin{itemize}
    \item We adapt BYOL to learn representations for skeleton-based action recognition.
    \item We propose a data augmentation strategy for skeleton sequences including asymmetric transformation pipelines and multi-viewpoint sampling. 
	\item We conduct extensive experiments on three large-scale datasets for skeleton-based action recognition, NTU-60, NTU-120 and PKU-MMD, to demonstrate the performance of our proposed method. Our method consistently outperforms the current state of the art on linear evaluation, semi-supervised and transfer learning benchmarks.

\end{itemize}

\section{Related Work}

{\flushleft \bf Skeleton-based Action recognition.}
Early skeleton-based action recognition methods relied on hand-crafted features \cite{Xia2012-bh, Wang2014-hg, Vemulapalli2014-ei}.
RNN-based methods treat skeleton data as vector sequences \cite{Du2015-by, Shahroudy2016-xf, Song2017-im, Zhang2017-jy}, while CNN approaches transform the skeleton data into image-like formats amenable to convolutions \cite{Ke2017-wh, Liu2017-gk, Kim2017-ba, Li2017-ya, Li2017-sm, Li2018-sa}.

Yan et al.\ \cite{Yan2018-pk} introduced a graph structure called Spatial-Temporal Graph Convolutional Network (ST-GCN), in which spatial edges connect body joints and temporal edges connect each joint across time steps. 
Many recent methods adopt the ST-GCN architecture \cite{Li2019-ml, Liu2020-pl, Li2019-uf, Shi2019-wc, Shi2019-yj, Peng2020-jd, Cheng2020-wl, Song2020-ly} and enhance it with new features such as adaptive adjacency, self-attention, and ensembling of multiple input streams.

{\flushleft \bf Self-supervised learning.}
A common approach to representation learning is to train a network on a large, labeled dataset. 
Self-supervised learning aims at learning transferable representations without using labels by extracting supervisory signals from the data itself.
Early methods relied on pretext tasks related to high-level image understanding, 
including reconstruction tasks such as denoising \cite{Vincent2008-eh}, inpainting \cite{Pathak2016-qc} or colorizing \cite{Zhang2016-th, Larsson2016-hr, Zhang2017-lj}, 
or transformation prediction tasks such as jigsaw puzzles \cite{Doersch2015-of, Noroozi2016-ns} or image rotations \cite{Gidaris2018-wh}.
Discriminative methods based on contrastive learning \cite{Hadsell2006-yi,Van_den_Oord2018-ms,Misra2020-tg,Chen2020-wj,Chen2020-bo,He2020-nt, Chen2020-tw} have recently achieved new state-of-the-art results on linear evaluation and semi-supervised learning tasks. Moreover, when transferring the learned representation to downstream tasks such as object detection, they also demonstrated performance
on par with or surpassing supervised pre-training. These methods learn by constructing positive and negative sample pairs and training the network to generate representations that can be used to distinguish the positive pairs from the negative ones.
A large number of negative samples is usually needed to achieve the best performance. For example, SimCLR \cite{Chen2020-wj, Chen2020-bo} uses large batch sizes, and MoCo \cite{He2020-nt, Chen2020-tw} keeps a large queue of past representations as negative samples.

Several recent works showed that useful representations can be learnt without the need for explicit negative samples \cite{Grill2020-ey, Chen2020-vj, Caron2021-tp}.
BYOL \cite{Grill2020-ey}, in particular, trains an \emph{online} network to predict the representation generated by a \emph{target} network given two augmented views of the same input. 
Our work adapts BYOL for skeleton-based action recognition. While BYOL uses two slightly different augmentation sets, we use two very distinct conservative and aggressive augmentation pipelines, and we leverage recordings of the same action captured simultaneously from different camera angles to encourage learning view-invariant representations.

{\flushleft \bf Self-supervision for Skeleton-based Action Recognition.}
Self-supervised methods for learning action representations from skeleton sequences focused initially on solving pretext tasks.
Zheng \etal \cite{Zheng2018-el} used a recurrent encoder-decoder GAN to learn representations by reconstructing masked sequences.
Predict \& Cluster \cite{Su2020-st} trained a sequential auto-encoder to predict skeleton motion while keeping the decoder's weights fixed to force the encoder to learn discriminative latent representations.
EnGAN \cite{Kundu2018-ep} used a variational auto-encoder to learn a pose embedding manifold, and an additional encoder-decoder to learn action representations as trajectories on the pose manifold.
Later works adopted the contrastive learning paradigm.
MS$^2$L \cite{Lin2020-rf} integrated contrastive learning in a multi-task learning framework, 
and AS-CAL \cite{Rao2021-wk} used different augmentations of skeleton sequences to produce positive and negative pairs.
Thoker \etal \cite{Thoker2021-hc} performed representation learning with two different network architectures operating on graph-based and sequence-based modalities, in a cross-contrastive manner.
Recently, SkeletonCLR \cite{Li2021-rh} learned skeleton sequence representations with a momentum contrast framework, while CrosSCLR \cite{Li2021-rh} trained three networks using different modalities (joints, bones, motion) and mined similar sequences between modalities. 
 In a concurrent work, AimCLR \cite{guo2022aimclr} builds on SkeletonCLR, adding an energy-based attention-guided drop module and nearest neighbours mining. They also use normal and extreme augmentation pipelines, and propose to minimize the distributional divergence between the normally augmented and the extremely augmented views.

\section{Method}
In this section, we first review BYOL, a state-of-the-art self-supervised representation learning algorithm upon which our method is based. We then describe our data augmentation and multi-viewpoint sampling strategies. Finally, we describe the steps for knowledge distillation.

\subsection{Bootstrap Your Own Latent (BYOL)}\label{sec:method-ssl}

As shown in ~\cref{fig:byol}, BYOL consists of two neural networks. 
The \textit{target} network, defined by weights $\xi$, comprises an encoder $f_{\xi}$ and a projector $g_{\xi}$. 
The \textit{online} network is defined by parameters $\theta$ and consists of an encoder $f_{\theta}$, a projector $g_{\theta}$ and a predictor $q_{\theta}$.
BYOL performs representation learning by maximizing the similarity between the embeddings of two augmentations of the same action sequence generated by these networks.

More specifically, given an action sequence $\textbf{s}$, two random augmentations $\mathbf{v_{1}} = t_{1}(\mathbf{s})$ and $\mathbf{v_2} = t_{2}(\mathbf{s})$ are generated using $t_{1} \sim \mathcal{T}_{1}, t_{2} \sim \mathcal{T}_{2}$, where $\mathcal{T}_{1}$ and $\mathcal{T}_{2}$ denote two sets of transformations.
The view $\mathbf{v_1}$ is fed to the online network, where the encoder produces the representation $\mathbf{y_1} = f_{\theta}(\mathbf{v_1})$. This is then projected to an 
embedding $\mathbf{z_1} = g_{\theta}(\mathbf{y_1})$, which is fed to the predictor, yielding $q_{\theta}(\mathbf{z_1})$. 
Likewise, the view $\mathbf{v_{2}}$ is fed to the target network, which produces the representation $\mathbf{y_{2}} = f_{\xi}(\mathbf{v_{2}})$ and projection $\mathbf{z_{2}} = g_{\xi}(\mathbf{y_{2}})$.
Given $\mathbf{z_{1}}$, the prediction network $q_{\theta}$ is trained to predict the target $\mathbf{z_{2}}$ by minimizing the normalized $\ell_2$ distance
\begin{equation}
\mathcal{L}_{\theta,\xi}(\mathbf{v_{1}},\mathbf{v_{2}}) = 
\left\|\overline{q_\theta(\mathbf{z_{1}})} - \overline{\mathbf{z_{2}}}\right\|^{2}_{2}\,,
\label{eq:eq-byol-loss}
\end{equation}  
where $\overline{\mathbf{u}} = \mathbf{u}/\|\mathbf{u}\|_2$.
The loss is symmetrized by also inputting $\mathbf{v_{2}}$ to the online network and $\mathbf{v_{1}}$ to the target network and calculating $\mathcal{L}_{\theta,\xi}(\mathbf{v_{2}},\mathbf{v_{1}})$.
This gives the symmetric loss 
\begin{equation}
\mathcal{L}_{\theta,\xi}^{\mathrm{BYOL}} = \mathcal{L}_{\theta,\xi}(\mathbf{v_{1}},\mathbf{v_{2}}) +  \mathcal{L}_{\theta,\xi}(\mathbf{v_{2}},\mathbf{v_{1}})\,.
\label{eq:eq-sym-byol-loss}
\end{equation}
At each iteration, we optimize the loss in \cref{eq:eq-sym-byol-loss} only with regard to the online weights $\theta$, while the target network's weights $\xi$ are kept fixed.
The target network is updated from past iterations of the online network, using an exponential moving average (EMA) on the weights $\theta$ after each training step. i.e.
\begin{equation}
\xi \leftarrow \lambda\xi + (1 - \lambda)\theta, \ \lambda \in [0,1]\,.
\end{equation}

{\flushleft \bf Momentum BatchNorm.}
BYOL performs best when calculating stable batch statistics over large mini-batches across GPUs and compute nodes. This synchronized batch normalization (SyncBN) relies on cross-GPU communication and leads to slower training speeds.
As the target network does not propagate gradients, Cai \etal \cite{Cai2021-jp} and Li, Liu and Sun \cite{Li2021-ke} showed that the need for SyncBN can be eliminated by employing momentum BN layers in the target network, which keep a moving average of the batch statistics.
After each training iteration, the mean $\mu$ and variance $\sigma$ of every momentum BN layer in the target network is updated using an exponential moving average of the current batch statistics of the online network, i.e.
\begin{equation}
\begin{aligned}
\mu \leftarrow \alpha\mu + (1 - \alpha)\mu_{b}\,,\quad
\sigma \leftarrow \alpha\sigma + (1 - \alpha)\sigma_{b}\,,
\end{aligned}
\end{equation}
where $\alpha \in [0,1]$ is the momentum coefficient, $\mu_{b}$ and $\sigma_{b}$ are the current batch statistics.
We find that using momentum BN instead of SyncBN reduces training times by 30\%.

\subsection{Data Augmentation}\label{sec:method-aug}
As the encoder learns to map augmented views of the same skeleton sequence close to each other in latent space, it is trained to ignore the variances induced by the augmentations.
The choice of data augmentation is thus critical for learning representations that are semantically relevant for action recognition.
We use the following augmentation strategies:

{\flushleft \bf Shear.}
Shearing \cite{Rao2021-wk} is a linear transformation that slants the shape of an object in a given direction. 
The shear transformation matrix is defined as
\begin{equation} \boldsymbol{S} =\left[\begin{array}{ccc}
1 & {s}_{X}^{Y} & {s}_{X}^{Z} \\
{s}_{Y}^{X} & 1 & {s}_{Y}^{Z} \\
{s}_{Z}^{X} & {s}_{Z}^{Y} & 1
\end{array}\right], \end{equation}
where $s_{X}^{Y}, s_{X}^{Z}, s_{Y}^{X}, s_{Y}^{Z}, s_{Z}^{X}, s_{Z}^{Y}$ are the shear factors defining the displacement direction.
We apply the transformation by multiplying the skeleton sequence with $S$ on the channel dimension.

{\flushleft \bf Left/Right Drop.}
This transformation randomly zeroes out the left or right limbs of the skeleton during the entire sequence
so the network learns to extract as much information as possible from partial observations of the body.

{\flushleft \bf Resampling.}
Leveraging the fact that the same action can be performed at various speeds, we speed up or slow down the input sequence by resampling it along the temporal dimension according to a random rate.

{\flushleft \bf Filtering.}
Skeleton sequences captured by Kinect cameras are quite noisy, which is apparent in the NTU-60 and NTU-120 datasets. We train the network to be robust to noise by randomly smoothing the sequences with a low-pass filter.

{\flushleft \bf Temporal shift.}
In real-world applications, action recognition algorithms are typically applied on overlapping temporal windows, which means that the action of interest might have already started, or might start or end anytime during a window. We therefore shift the sequence cyclically along the time dimension with a random offset.

\begin{figure}[t]
\centering
\includegraphics[width=0.8\linewidth]{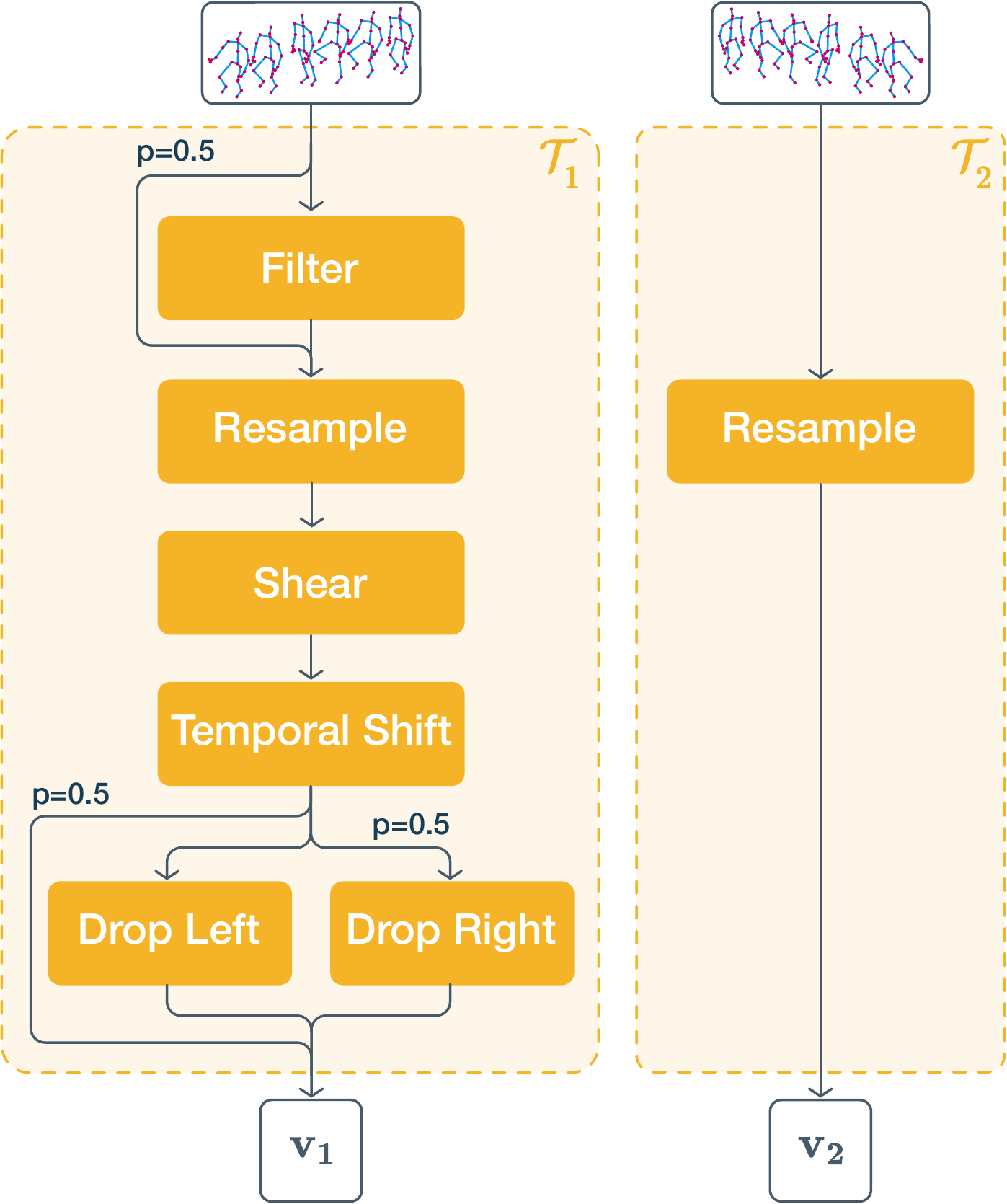}
\caption{\label{fig:asym}\textbf{Asymmetric Augmentation Pipelines.}
 }
\vspace{-5pt} 
\end{figure}

\subsection{Asymmetric augmentation pipelines} 
While strong augmentation has been shown to be beneficial for self-supervised learning \cite{Grill2020-ey}, some of the augmentations above heavily distort the augmented samples, which may become very different from the original data. 
By contrast, overly aggressive augmentations empirically lead to worse results during supervised fine-tuning, therefore only conservative augmentations are used in that setting.
To enforce consistency between the representations of the data seen during self-supervised pre-training and in the fine-tuning stage, we use two very different transformation distributions during self-supervised learning.

In the \textit{aggressive} distribution $\mathcal{T}_1$, filtering, random resampling, shear, temporal shift and left/right drop are available. Filtering and left/right drop are randomly activated, and each side of the body has equal probability of being dropped.
In the \textit{conservative} distribution $\mathcal{T}_2$, only random resampling is enabled.
The augmentation pipelines are illustrated in \cref{fig:asym}.

Our experiments show that networks pre-trained with the asymmetric pipelines achieve better performance on downstream tasks.

\subsection{Multi-Viewpoint Sampling}
It is common for skeleton sequence datasets to contain recordings of the same action sequences captured simultaneously from different angles by different cameras.
Although the 3D skeletal representation provided by RGB-D cameras such as Kinect is in essence view-invariant, it is still sensitive to errors of the body tracker, due \eg to self-occlusions \cite{Shahroudy2016-xf}.
Unlike action labels, viewpoint information is free to obtain as it is provided by the capturing system.
We propose to leverage this information to learn representations that are invariant to changes of viewpoint and to the errors of the pose algorithm.

Let $\mathcal{S}$ be the dataset. Each skeleton sequence $\mathbf{s_{i,c}} \in \mathcal{S}$ represents a viewpoint of the action sequence $\mathbf{s_{i}}$ captured by camera $c \in \{1..C\}$.
Under multi-viewpoint sampling, instead of iterating over the skeleton sequences $\mathbf{s_{i,c}}$, we iterate over the action sequences $\mathbf{s_{i}} = \{\mathbf{s_{i,1}},...,\mathbf{s_{i,C}}\}$. When $\mathbf{s_{i}}$ contains more than one viewpoint, we draw two random viewpoints $\mathbf{s_{i,c_1}}$, $\mathbf{s_{i,c_2}}$ from $\mathbf{s_{i}}$, with replacement (i.e., the two viewpoints may be identical), to construct a sequence pair.
The two viewpoints are passed to the transformations $t_{1} \sim \mathcal{T}_{1}, t_{2} \sim \mathcal{T}_{2}$ to form $\mathbf{v_{1}} = t_{1}(\mathbf{s_{i,c_1}})$ and $\mathbf{v_2} = t_{2}(\mathbf{s_{i,c_2}})$, respectively.
We find that multi-viewpoint sampling significantly improves the accuracy of networks trained with our method.

\subsection{Knowledge Distillation}\label{sec:method-distill}
For self-supervised learning, it is beneficial to use larger, wider networks. This, however, increases the cost for inference. Inspired by \cite{Chen2020-bo}, we use knowledge distillation \cite{Hinton2015-pj} to transfer the knowledge from a large model to a smaller model to address this problem. Given a fixed \textit{teacher network} (fine-tuned on a fraction of the labels), we train a randomly-initialized \textit{student network} to match the output predictions of the teacher by minimizing the loss function
\begin{equation}
L = H_{CE}\left(\sigma\left(\frac{z_t(x)}{\tau}\right), \sigma\left(\frac{z_s(x)}{\tau}\right)\right),
\end{equation}
                                            
where $H_{CE}(.)$ is the cross-entropy loss, $\sigma(.)$ is the softmax function, $z_s(x)$ and $z_t(x)$ are the output logits of the student and teacher networks respectively, and $\tau$ is a temperature parameter controlling the softness of the teacher's output. Note that this step does not require any labeled data.

\section{Experiments}
We follow standard practice and evaluate our representations under the linear evaluation and semi-supervised protocols.
We also demonstrate the performance of our method on a transfer learning task.
We perform an ablation study to illustrate the effectiveness of each component in our method, and conduct comprehensive comparisons with existing state-of-the-art methods.

\subsection{Datasets}

{\flushleft \bf NTU-60.}
NTU-60 \cite{Shahroudy2016-xf} is a large-scale action-recognition dataset containing over 56K samples of 60 different actions performed by 40 subjects, recorded simultaneously from different viewpoints by Kinect v2 sensors.
We follow the splits recommended by the authors: in Cross-Subject (CS), the training and test sets each contains data from 20 different subjects, and in Cross-View (CV) the training set contains action sequences captured by cameras 2 and 3 while the test set contains the same sequences captured by camera 1.

{\flushleft \bf NTU-120.}
NTU-120 \cite{Liu2019-ri} is an extension of NTU-60 and contains 60 additional action classes.
It contains 114K samples of 120 actions performed by 106 different subjects.
The dataset contains 32 setups denoting specific locations and backgrounds.
We follow the standard splits defined by the authors: for Cross-Subject (CSub), the training and test sets contain data from 53 subjects each, and for Cross-Setup (CSet) the training set contains data from all samples with even setup IDs, and the test set those with odd setup IDs.

{\flushleft \bf PKU-MMD II.}
PKU-MMD II \cite{liu2017pku} 
is a large-scale multi-modal dataset for continuous 3D human action detection, 
consisting of 2000 short sequences containing 7K instances of 41 action classes performed by 13 subjects and recorded from three viewpoints by Kinect v2 cameras.
We follow the Cross-Subject split provided by the authors.

\subsection{Implementation details}\label{sec-impl-detail}

{\flushleft \bf Encoder network and data representation.}
We use the original ST-GCN \cite{Yan2018-pk}, both as the encoder in our framework and as a strong supervised baseline for our experiments. 
In experiments with encoders of different widths, we multiply the number of channels in all layers.
For example, a $2 \times$ network has twice the number of channels in each layer compared with the standard
ST-GCN, whereas a $0.5 \times$ network has half as many.

In addition to the 3D joint coordinates, we use the bone representation, calculated as the vectors between the two vertices of each edge of the skeleton graph \cite{Shi2019-wc}.
Several recent action recognition methods use multiple modalities, including
joint coordinates and bone representation, either by ensembling different models or by employing multi-stream architectures \cite{Shi2019-wc, Song2020-ly}.
Here, we simply concatenate the joint and bone representations along the channel dimension. 
This only affects the input layer of the network and does not require any further modification to the ST-GCN architecture.

{\flushleft \bf Data processing.}
All skeleton sequences are cropped to 150 frames, and sequences shorter than 150 frames are repeated.
The skeletons are centered, and rotated so that the body has a canonical orientation in the first frame.

{\flushleft \bf Data augmentation.}
In the \textit{aggressive} augmentation pipeline, we resample the skeleton sequences by a rate uniformly sampled in the interval $[0.7, 1.3]$. Filtering is applied randomly with a probability of $0.5$, and is performed using a Savitzky-Golay filter with order 2 and window 15. The shear factors are sampled randomly from $[-1, 1]$. The sequence is cyclically shifted along the temporal dimension with an offset uniformly sampled in the interval $[0,150]$. The left/right drop augmentation is applied with a probability of $0.5$, and each side of the body has equal probability of being dropped.

In the \textit{conservative} augmentation pipeline, only random resampling is performed.

{\flushleft \bf Supervised baseline training.}
We train the fully supervised model for 80 epochs with a mini-batch size of 8 on a single GPU.
We use SGD with weight decay 0.0003 and momentum 0.9.
The initial learning rate is 0.0125, and is multiplied by 0.1 at epochs 56 and 70.

{\flushleft \bf Self-supervised pre-training.} 
During self-supervised pre-training, we train the model for 1600 epochs with a mini-batch size of 512 on 4 V100 GPUs, which takes approximately 16 hours. 
For the much smaller PKU-MMD II dataset, we train for 4800 epochs.
For wider ($2\times$) networks, pre-training takes 23 hours on 8 V100 GPUs.
We use SGD with weight decay $10^{-4}$ and momentum 0.9. 
The learning rate starts at $10^{-6}$ and is linearly increased to 0.2 in the first 10 epochs of training, after which it is decreased to 0 following a cosine decay schedule \cite{Loshchilov2016-mf}. 
We initially set the BYOL momentum $\lambda$ to 0.99 and increase it to 1 following a cosine schedule during training.

\subsection{Linear Classification Protocol}
\label{sec:lineval}


\begin{table}[t]
\begin{adjustbox}{width=1\linewidth}
\small
\centering
\setlength{\tabcolsep}{5pt} 
\begin{tabular}[t]{l cc cc c}
\toprule[1pt]
                      &\multicolumn{2}{c}{NTU-60}&\multicolumn{2}{c}{NTU-120}&PKU-MMD II\T\\                    
 Method             & CS & CV & CSub & CSet & CS\T\B\\
\midrule[.5pt]
ST-GCN (sup.)                  & \textbf{88.5}     & \textbf{94.3} & \textbf{83.0}     & \textbf{85.1}& \textbf{61.9}\T\B\\
\midrule[.5pt]
LongT GAN \cite{Zheng2018-el}                  & 39.1 & 48.1  & - & - & 26.5\T\\
MS$^{2}$L \cite{Lin2020-rf}                        & 52.6 & - & -& - & 27.6\\
PCRP \cite{Xu2020-cp}                        & 53.9 & 63.5 & -& - & -\\
AS-CAL \cite{Rao2021-wk}                        & 58.5 & 64.8 & 48.6 & 49.2  & -\\
Thoker \etal \cite{Thoker2021-hc}                       & 76.3 & 85.2 & 67.1 & 67.9 & 36.0\\
3s-CrosSCLR \cite{Li2021-rh}                       & 77.8 & 83.4 & 67.9 & 66.7 & -\\
3s-AimCLR \cite{guo2022aimclr}                 & 78.9 & 83.8 & 68.2 & 68.8 & -\B\\
\midrule[.5pt]
Ours                         &  \textbf{86.8} &  \textbf{91.2}  &  \textbf{77.1} &  \textbf{79.2} & \textbf{55.25}\T\B\\
\bottomrule[1pt]
\end{tabular}
\end{adjustbox}
\caption{\label{table:lineareval}\textbf{Linear evaluation protocol.} Performance comparison (Top-1 accuracy, \%) with the fully-supervised baseline and other self-supervised methods on NTU-60, NTU-120 and PKU-MMD.
}
\vspace{-5pt}
\end{table}


We first evaluate the quality of the representations learned with our method under the linear classification protocol,
training a linear classifier on top of the frozen pre-trained online encoder $f_{\theta}$.
We train for 100 epochs with a batch size of 256 on 4 V100 GPUs, using SGD with momentum 0.9 and without weight decay.
The initial learning rate is 30, and is multiplied by 0.1 at epochs 60 and 80.
We only use conservative augmentations in this step.

\begin{table*}[t]
\small
\centering
\setlength{\tabcolsep}{9pt}
\begin{tabular}{l cc l cc}
\toprule[1pt]
Method  & MVS & Asymm. Aug. & Augmentations & NTU-60 (CS) & NTU-60 (CV) \T\B\\
\midrule[.5pt]
Baseline BYOL & & & Only conservative augmentation  & 51.1 & 55.1 \T\\
 & & & Only aggressive augmentation  & 79.6 & 81.0\B\\
\midrule[.5pt]
Ours & $\checkmark$ & $\checkmark$ & No shear          &  78.0 & 81.1\T\\
& $\checkmark$ & $\checkmark$ & No low-pass filtering      &  84.4 & 90.2\\
& $\checkmark$ & $\checkmark$ & No temporal shift         &  84.7 & 89.4\\
& $\checkmark$ & $\checkmark$ & No resampling     &  85.6 & 90.6\\
& $\checkmark$ & $\checkmark$ & No left/right drop        &  85.6 & 89.7\B\\
\cmidrule(lr{1em}){2-6}
& & $\checkmark$ & All augmentations     &  81.8 & 82.3\T\\
& $\checkmark$ & & Only aggressive augmentation &  83.8  & 88.7\B\\
\cmidrule(lr{1em}){2-6}
& $\checkmark$ & $\checkmark$ & All augmentations &  \textbf{86.8} & \textbf{91.2} \T\B\\
\bottomrule[1pt]
\end{tabular}
\caption{\label{table:ablation}\textbf{Ablation study} on the components of our self-supervised training method. We remove multi-viewpoint sampling (MVS), the asymmetric augmentation pipelines (Asymm. Aug.) and each augmentation in turn. Linear evaluation (Top-1 accuracy, \%) on NTU-60.
}
\vspace{-5pt}
\end{table*}

{\flushleft \bf Comparison with previous results}
We compare our method with the fully-supervised baseline and with other self-supervised methods by linear evaluation on the NTU-60, NTU-120 and PKU-MMD II datasets.
As shown in \cref{table:lineareval}, \OURS outperforms all other self-supervised methods and considerably reduces the gap to fully-supervised training.

{\flushleft \bf Ablation Studies}

We evaluate the usefulness of the components of our method by removing them in turn.
Note that removing both multi-viewpoint sampling and the asymmetric augmentation pipelines corresponds to applying the classical BYOL method to skeleton sequence data.

The results are shown in \cref{table:ablation}.
They demonstrate the performance of the aggressive augmentation pipeline, and show that all data augmentations contribute to the method's performance, with shearing being the most effective transformation.
The results also demonstrate the performance gains obtained with multi-viewpoint sampling and with the use of two asymmetric augmentation pipelines.
In the cross-view setting, the shear transformation and multi-viewpoint sampling both provide the most significant performance increase, reinforcing our claim that these features force the network to learn invariances to change of viewpoint.
Finally, for both benchmarks, the combination of all our contributions provides the best results.

\begin{figure}
\centering
\begin{tabular}{cccc}
\includegraphics[width=.45\linewidth]{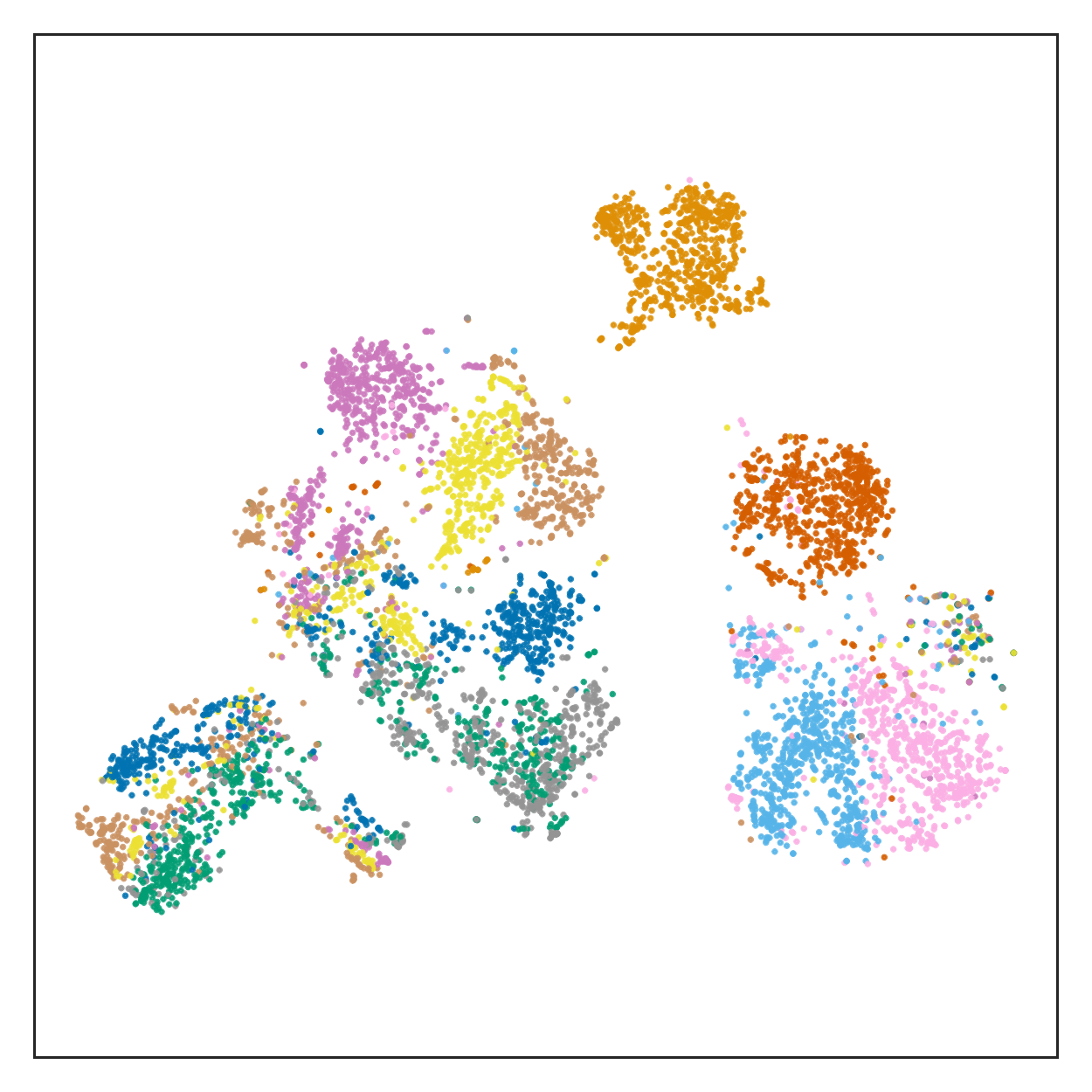}&
\includegraphics[width=.45\linewidth]{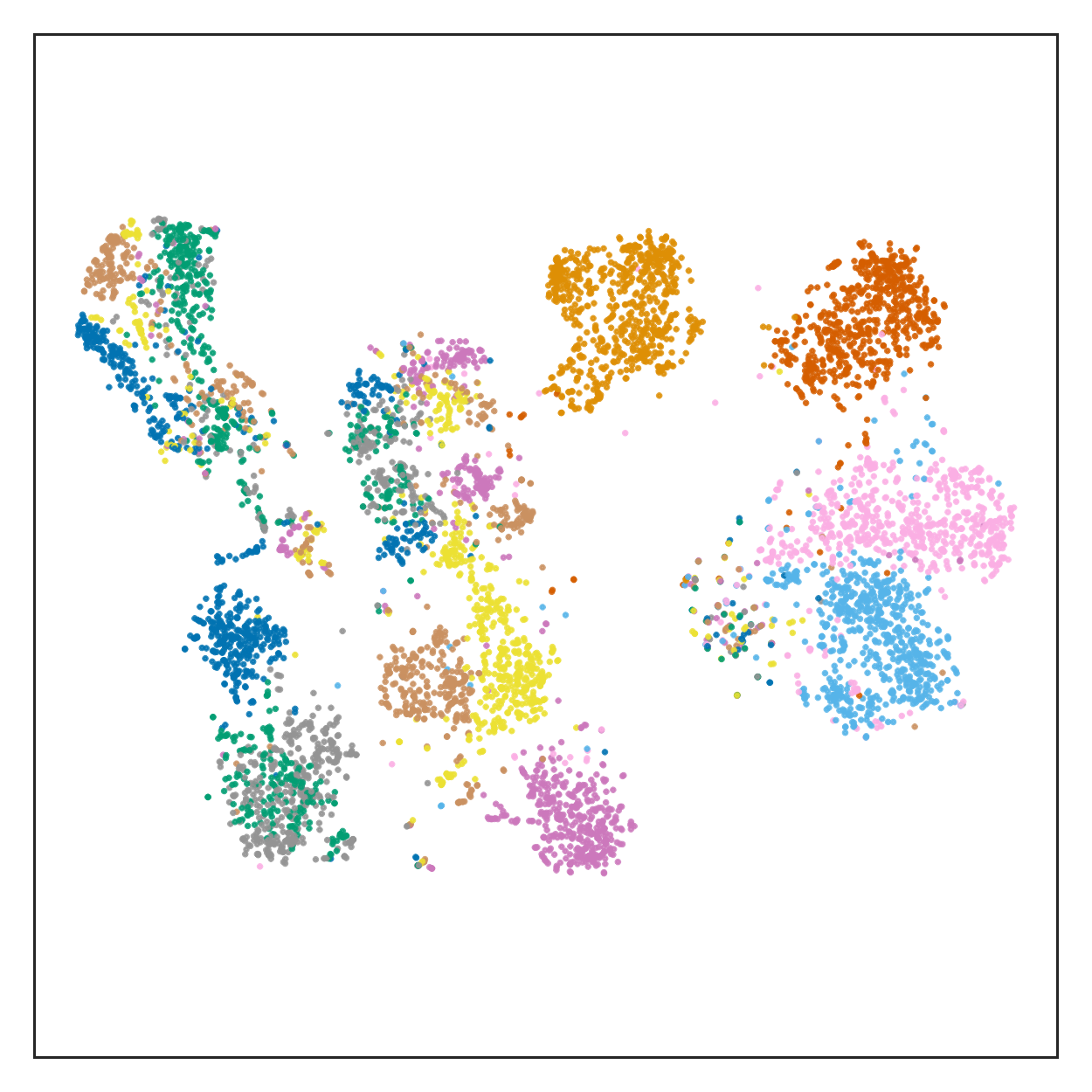} \\
(a)&(b)\\
\includegraphics[width=.45\linewidth]{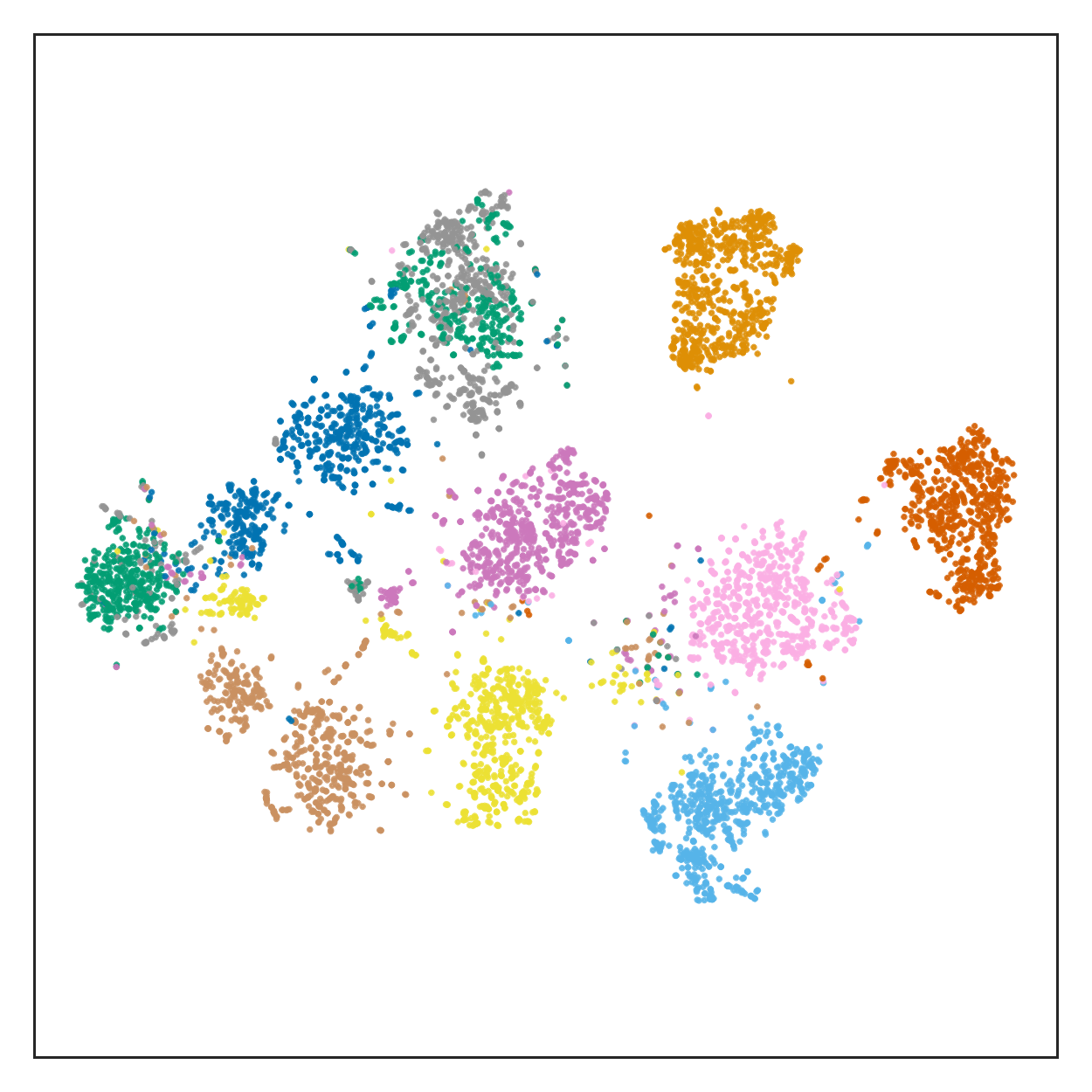}&
\includegraphics[width=.45\linewidth]{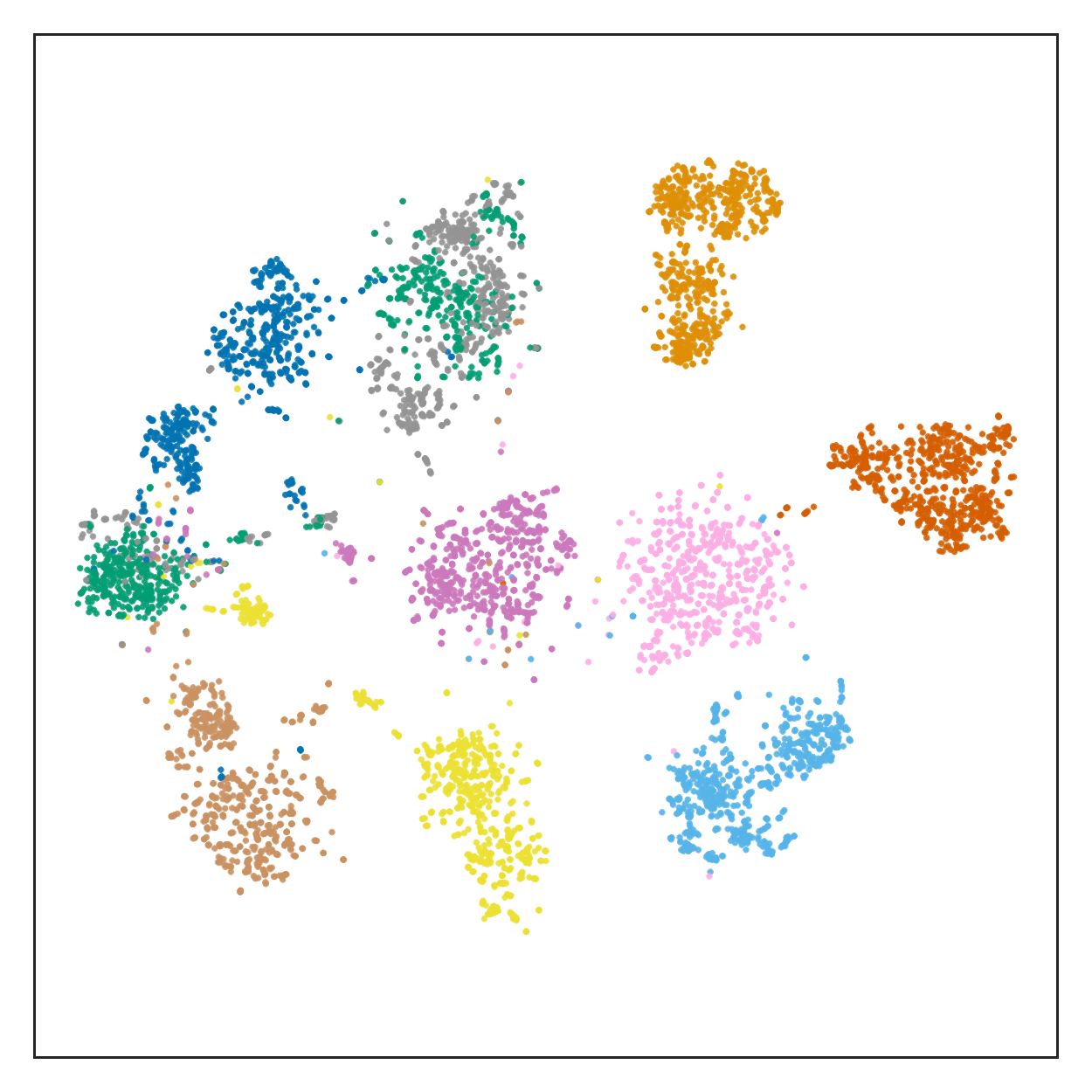}\\
(c)&(d)
\end{tabular}
\caption{\textbf{t-SNE visualization of the features} learned under self-supervised training with a) Baseline BYOL with aggressive augmentations, b) Our method without multi-viewpoint learning, c) Our method with only aggressive augmentations and d) Our method. Each point represents a skeleton sequence. We show 10 random action classes of the NTU-60 dataset, indicated by colors.}
\label{fig:tsne}
\vspace{-5pt}
\end{figure}

{\flushleft \bf Visualization of Learned Representations}
We present in \cref{fig:tsne} a t-SNE \cite{JMLR:v9:vandermaaten08a} visualization of the representations learned under self-supervised training with the BYOL baseline and with our proposed method, with and without multi-viewpoint learning and asymmetric augmentation pipelines.
Our proposed multi-viewpoint learning and asymmetric augmentation pipelines lead to a better separation of the action classes.

\subsection{Semi-Supervised Evaluation}
\label{sec:semisup}
We now evaluate our learned representation on semi-supervised learning tasks.
We sample 1\%, 5\% and 10\% of the NTU-60 training set in a class-balanced way as labeled samples and fine-tune the encoder $f_{\theta}(.)$ 
with a linear layer on top. 
We report top-1 accuracy on the entire test set. We present results averaged over five randomly sampled subsets of the labeled data.

When fine-tuning the network on 1\%, 5\% and 10\% of the labeled samples, we train for 100, 20 and 10 epochs respectively, with a mini-batch size of 8 on a single Nvidia V100 GPU using SGD with momentum 0.9 and without weight decay.
The initial learning rate is 10, and is decayed to 0 using a cosine schedule.
We keep the pre-trained encoder frozen during the first third of the iterations.
Afterwards, the encoder's learning rate follows the same schedule as the fully connected layer's but is multiplied by a factor of $10^{-3}$.
Only conservative augmentations are used.

We then distill the fine-tuned model to another network using the whole training set \textit{without labels}.
We train the student network for 100 epochs with a batch size of 256, using SGD with weight decay $10^{-4}$, momentum 0.9, and distillation temperature 1.0.
The initial learning rate is 0.2, and is decayed to 0 with a cosine schedule.

\begin{table}[t]
\small
\centering
\setlength{\tabcolsep}{4pt} 
\begin{tabular}{l ccc!{\quad}ccc}
\toprule[1pt]
          &\multicolumn{3}{c}{NTU-60 (CS)}&\multicolumn{3}{c}{NTU-60 (CV)}\T\B\\
Label fraction &        1\% & 5\% & 10\%&        1\% & 5\% & 10\% \B\\
\midrule[.6pt]
ST-GCN (sup.)                  & 19.3 & 59.1 & 71.7 & 20.6 & 61.6 & 75.9 \T\B\\
\midrule[.6pt]
SESAR-KT \cite{Li2020-pm}                   & 48.1 & 55.0 & 58.2 & - & - & - \T\\
MS$^{2}$L  \cite{Lin2020-rf}                   & 33.1 & - & 65.2 & - & - & - \\
ASSL \cite{Si2020-lx}            & - & 57.3 & 64.3 & - &  63.6 & 69.8\\
3s-CrosSCLR \cite{Li2021-rh}            & 51.1 & - & 74.4 & 50.0 & - & 77.8\\
Thoker \etal \cite{Thoker2021-hc}            & 35.7 & 59.6 & 65.9 & 38.1 & 65.7 & 72.5 \\
3s-AimCLR \cite{guo2022aimclr}            & 54.8 & - & 78.2 & 54.3 & - & 81.6\B\\
\midrule[.6pt]
Ours $1\times$, distilled           &  \textbf{79.4} &  83.6 &  84.6  & \textbf{81.7} & \textbf{87.5} & 89.5\T\\
Ours $2\times$                      &  79.3 &  \textbf{84.5} &  \textbf{86.0} & 81.5 & \textbf{87.5} & \textbf{89.8} \B\\
\bottomrule[1pt]
\end{tabular}
\caption{\label{table:semisup} \textbf{Semi-supervised learning on NTU-60} with different label fractions for fine-tuning. The labels are sampled in a class-balanced way. For our method, we present results averaged over five random seeds for sampling the labeled data.
}
\vspace{-6pt}
\end{table}

\begin{table}[t]
\small
\centering
\setlength{\tabcolsep}{4pt} 
\begin{tabular}{l ccc!{\quad}ccc}
\toprule[1pt]
          &\multicolumn{3}{c}{NTU-120 (CSub)}&\multicolumn{3}{c}{NTU-120 (CSet)} \T\B\\
Label fraction &        1\% & 5\% & 10\%&        1\% & 5\% & 10\%\B\\
\midrule[.6pt]
ST-GCN (sup.)                  & 15.2 & 51.2 & 63.0 & 14.0 & 49.9 & 61.9 \T\B\\
\midrule[.6pt]
Ours $1\times$, distilled           & \textbf{66.8} & \textbf{74.2} & \textbf{77.0} & \textbf{67.1} & \textbf{76.2} & 78.0 \T\\
Ours $2\times$                      & 65.9 & 73.9 & 76.7 & 66.6 & 76.1 & \textbf{78.6} \B\\
\bottomrule[1pt]
\end{tabular}
\caption{\label{table:semisup120} \textbf{Semi-supervised learning on NTU-120} with different label fractions for fine-tuning. 
}
\vspace{-6pt}
\end{table}

{\flushleft \bf Comparison with previous results}
In \cref{table:semisup,table:semisup120} we compare our method with the fully-supervised baseline and other semi-supervised methods.
The results further highlight the quality of the action representations learned by our approach: fine-tuning the model on 1\% ($\sim$400) labeled sequences of NTU-60 (CS) yields a top-1 accuracy of 79.4\%, outperforming both the fully-supervised baseline and other semi-supervised methods by a large margin.

\begin{table}[t]
\small
\centering
\begin{tabular}{ll ccc}
\toprule[1pt]
          & &\multicolumn{3}{c}{Label fraction} \T\\
Model &    KD   & 1\% & 5\% & 10\% \B\\
\midrule[.5pt]
\multirow{2}{*}{Ours $0.25\times$}  &                      & 53.5 \scriptsize{$\pm$1.5} & 68.9 \scriptsize{$\pm$0.4} & 71.2 \scriptsize{$\pm$0.3} \T \\

 & $\checkmark$         &  76.8 \scriptsize{$\pm$0.7} &  80.4 \scriptsize{$\pm$1.4} &  81.3 \scriptsize{$\pm$0.9} \B\\
\midrule[.5pt]
\multirow{2}{*}{Ours $0.5\times$}   &                     & 66.9 \scriptsize{$\pm$0.8} & 78.4 \scriptsize{$\pm$0.6} & 80.4 \scriptsize{$\pm$0.3} \T\\
 & $\checkmark$            &  78.3 \scriptsize{$\pm$0.3} &  81.7 \scriptsize{$\pm$0.7} &  82.8 \scriptsize{$\pm$0.4} \B\\
\midrule[.5pt]
\multirow{2}{*}{Ours $1\times$}   &                     & 75.8 \scriptsize{$\pm$1.0} & 82.8 \scriptsize{$\pm$0.6} & 84.5 \scriptsize{$\pm$0.4} \T\\
 & $\checkmark$                        &  79.4 \scriptsize{$\pm$0.6} &  83.6 \scriptsize{$\pm$0.7}  &  84.6 \scriptsize{$\pm$0.3}  \B\\
\midrule[.5pt]
Ours $2\times$  &                    & 79.3 \scriptsize{$\pm$0.6} & 84.5 \scriptsize{$\pm$0.4} & 86.0 \scriptsize{$\pm$0.1} \T\B\\
\bottomrule[1pt]
\end{tabular}
\caption{\label{table:distillation}\textbf{Effect of distillation.} Top-1 accuracy (\%) for semi-supervised and distilled models of varied sizes on different label fractions of the NTU-60 (CS) dataset. Knowledge distillation (KD) is performed with the whole \textit{unlabeled} training set, using the $2\times$ models fine-tuned on each label fraction as teachers. We present the mean and standard deviation of the results obtained over five random seeds for sampling the labeled data.
}
\end{table}

{\flushleft \bf Effect of distillation}
To evaluate the effect of knowledge distillation, we compare semi-supervised and distilled ST-GCN models of varied widths 
on the NTU-60 (CS) dataset.
The semi-supervised models are pre-trained on the whole unlabeled training set and fine-tuned on different label fractions of the training set, while the distilled models are trained using fine-tuned $2\times$ models as teachers.
The results are shown in \cref{table:distillation}.
We observe that knowledge distillation is effective and consistently provides generalization gains, especially at lower label fractions, and for lower-capacity students.

\subsection{Transfer Learning}

We evaluate the transfer learning performance of our method by pre-training the encoder $f_{\theta}(.)$ on NTU-60 and fine-tuning it with a linear layer on top on the PKU-MMD II dataset.
The results are shown in \cref{table:transfer}.
When fine-tuned, our pre-trained model significantly outperforms both previous work and the supervised baseline.

\begin{table}[t]
\small
\centering
\setlength{\tabcolsep}{9pt}
\begin{tabular}[t]{l c}
\toprule[1pt]
 Method                     & PKU-MMD II \T\B\\                    
\midrule[.5pt]
ST-GCN (supervised)                  & \textbf{61.9}\T\B\\
\midrule[.5pt]
LongT GAN \cite{Zheng2018-el}                & 44.8 \T\\
MS$^{2}$L \cite{Lin2020-rf}                  & 45.8\\
Thoker \etal \cite{Thoker2021-hc}            & 45.9 \\
\midrule[.5pt]
Ours                         &  \textbf{68.3} \T\B\\
\bottomrule[1pt]
\end{tabular}
\caption{\label{table:transfer}\textbf{Transfer learning from NTU-60 to PKU-MMD II.} Performance comparison (Top-1 accuracy, \%) when fine-tuning on PKU-MMD II a model pre-trained on NTU-60.}
\vspace{-5pt}
\end{table}

\section{Discussion}
\subsection{Limitations}
Not all datasets provide multiple camera views of the same action sequence, making it impossible to take advantage of our proposed multi-viewpoint sampling.
However, as seen in \cref{sec:lineval}, our method still provides competitive results without this contribution.

The benchmarks used by this and previous work are based on datasets in which the classes are balanced and the class distributions of the training and test sets are similar.
The results may not be representative of the performance obtained in real applications, where unlabeled datasets are often imbalanced, with long-tailed label distributions.
Investigating the robustness of self-supervised skeleton-based action representation learning to dataset imbalance requires further exploration, which we leave to future work.

\subsection{Potential Negative Societal Impact}
In contrast to fully-supervised learning, self-supervised representation learning requires training with large batches for a large number of epochs on power-intensive computing hardware, and has therefore a higher environmental impact.
However, as shown in \cref{sec:semisup}, knowledge distillation allows transferring the learned representations to smaller, more energy-efficient models with little accuracy loss, which may help offset this impact.

The motivations behind our work have been to decrease the cost and increase the ease of adoption of action recognition technologies.
Camera-based applications becoming pervasive in public spaces raises obvious privacy concerns.
Persons moving in monitored areas might not be aware of the presence of cameras, and may not have the possibility to express (or deny) consent.
In a long-term perspective, if the technology generalizes the omnipresence of cameras might affect individuals' sense of freedom, and could generate anxiety. 
Alleviating such concerns will require a considerable effort of transparency and pedagogy to inform about the presence and purpose of capturing devices, proactive security and data protection, and clear accountability.

Performing self-supervised representation learning on large unlabeled datasets has the potential to reduce human bias introduced when curating and annotating data. 
However, some groups that are underrepresented in society will be likewise underrepresented in the unlabeled data. 
Ensuring that models used to make decisions do not impact those groups unfairly will require sizable effort.

Finally, the applications we envisage are overwhelmingly positive, in domains such as healthcare, sports, entertainment or livestock farming. However, we are conscious that the same technology may be used 
for other purposes.

\section{Conclusion}

We presented a simple framework for self-supervised representation learning for skeleton-based action recognition based on BYOL.
It includes a data augmentation strategy for skeleton data based on two distinct transformation pipelines, and a multi-viewpoint learning method that makes better use of action sequences that are captured simultaneously by different cameras. 
 Experiments on three large scale action recognition datasets show that our method consistently outperforms the current state of  the art on linear evaluation, semi-supervised and transfer learning tasks.

\section*{Acknowledgment}
This work was partially supported by the Wallenberg AI, Autonomous Systems and Software Program (WASP) funded by the Knut and Alice Wallenberg Foundation.

{\small
\bibliographystyle{ieee_fullname}
\bibliography{egbib}
}

\end{document}